\documentclass[10pt,twocolumn,letterpaper]{article}

\usepackage[pagenumbers]{cvpr} 

\usepackage{tablefootnote}
\usepackage{booktabs}

\usepackage{colortbl}
\usepackage[ruled]{algorithm2e}
\newcommand{\dn}{\texttt{AnimeRig}}
\definecolor{cvprblue}{rgb}{0.21,0.49,0.74}
\usepackage[pagebackref,breaklinks,colorlinks,allcolors=cvprblue]{hyperref}

\def\paperID{1146} 
\def\confName{CVPR}
\def\confYear{2025}

\title{DRiVE: \underline{D}iffusion-based \underline{Ri}gging Empowers \\Generation of \underline{V}ersatile and \underline{E}xpressive Characters }
\def\authorBlock{
    Mingze Sun\textsuperscript{1}\textsuperscript{*} \quad 
    Junhao Chen\textsuperscript{1}\textsuperscript{*} \quad
    Junting Dong\textsuperscript{2}\textsuperscript{\textdagger} \quad
    Yurun Chen\textsuperscript{1} \quad
    Xinyu Jiang\textsuperscript{1} \quad
    Shiwei Mao\textsuperscript{1} \quad
    Puhua Jiang\textsuperscript{1} \quad
    Jingbo Wang\textsuperscript{2} \quad
    Bo Dai\textsuperscript{2} \quad
    Ruqi Huang\textsuperscript{1}\textsuperscript{\textdagger} \\
    
    \textsuperscript{1}Tsinghua Shenzhen International Graduate School, China \\
    \textsuperscript{2}Shanghai AI Laboratory, China  \\  
}

\begin{document}

\twocolumn[{

\renewcommand\twocolumn[1][]{#1}
\maketitle
\begin{center}
    \vspace{-4\baselineskip}
    \author{\authorBlock}
\end{center}

\begin{center}
    \captionsetup{type=figure}
    \centerline{\includegraphics[width=\linewidth]{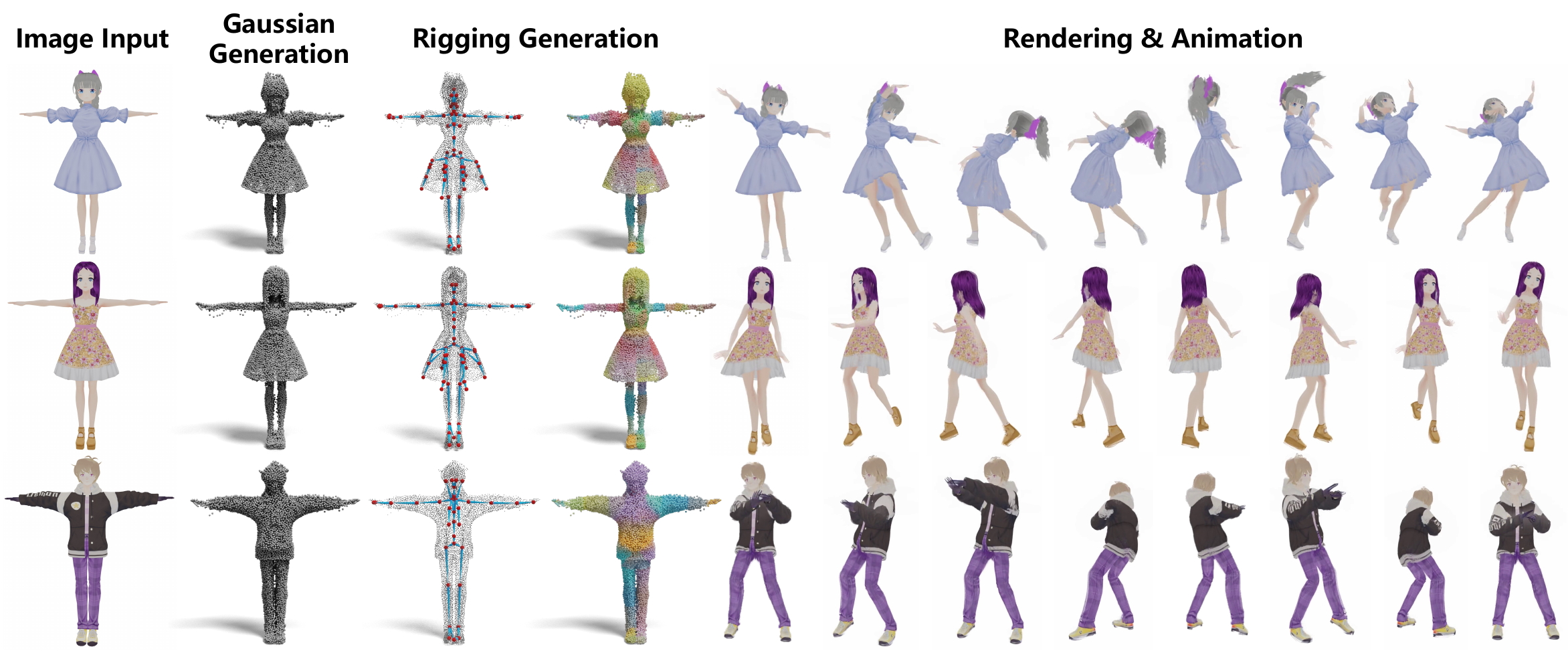}}
    \caption{
    We propose \textbf{DRiVE}, a pipeline that generates 3D Gaussian from a single image along with the corresponding skeleton (including hair and clothing) and skinning, enabling precise control over 3D Gaussian to render high-quality, controllable, and 3D consistent videos.
    }
    \label{fig:teaser}

\end{center}
}]

\renewcommand{\thefootnote}{} 
\footnotetext{\textsuperscript{*} Indicates Equal Contribution. \textsuperscript{\textdagger} Indicates Corresponding Author.}
\renewcommand{\thefootnote}{\arabic{footnote}} 

\begin{abstract}
Recent advances in generative models have enabled high-quality 3D character reconstruction from multi-modal. 
However, animating these generated characters remains a challenging task, especially for complex elements like garments and hair, due to the lack of large-scale datasets and effective rigging methods. 
To address this gap, we curate $\dn$, a large-scale dataset with detailed skeleton and skinning annotations. 
Building upon this, we propose \textbf{DRiVE}, a novel framework for generating and rigging 3D human characters with intricate structures. 
Unlike existing methods, DRiVE utilizes a 3D Gaussian representation, facilitating efficient animation and high-quality rendering. 
We further introduce GSDiff, a 3D Gaussian-based diffusion module that predicts joint positions as spatial distributions, overcoming the limitations of regression-based approaches. 
Extensive experiments demonstrate that DRiVE achieves precise rigging results, enabling realistic dynamics for clothing and hair, and surpassing previous methods in both quality and versatility.
The code and dataset will be made public for academic use at \href{https://DRiVEAvatar.github.io/}{https://DRiVEAvatar.github.io/}.

\end{abstract}    
\section{Introduction}\label{sec:intro}

Crafting and animating 3D human characters has long been a critical task in an array of applications, including film and video game making, AR/VR, and human-centric robotics, to name a few. 
Notably, the rapid development of generative models opens numerous new opportunities for research on this classic task. 
In contrast to the traditional manual and time-consuming crafting pipeline, nowadays one can \emph{create} high-quality graphical human models with \emph{rich structures} (\emph{e.g., }intricate garment and hairs) from a single image~\cite{saito2019pifu, saito2020pifuhd, xiu2022icon, xiu2023econ, peng2024charactergen}, a piece of text prompt~\cite{huang2024humannorm}, or a video clip~\cite{hu2023gaussianavatar, xiu2024puzzleavatar} automatically and efficiently. 

\begin{figure}[t!]
  \centering
  \includegraphics[width=\linewidth]{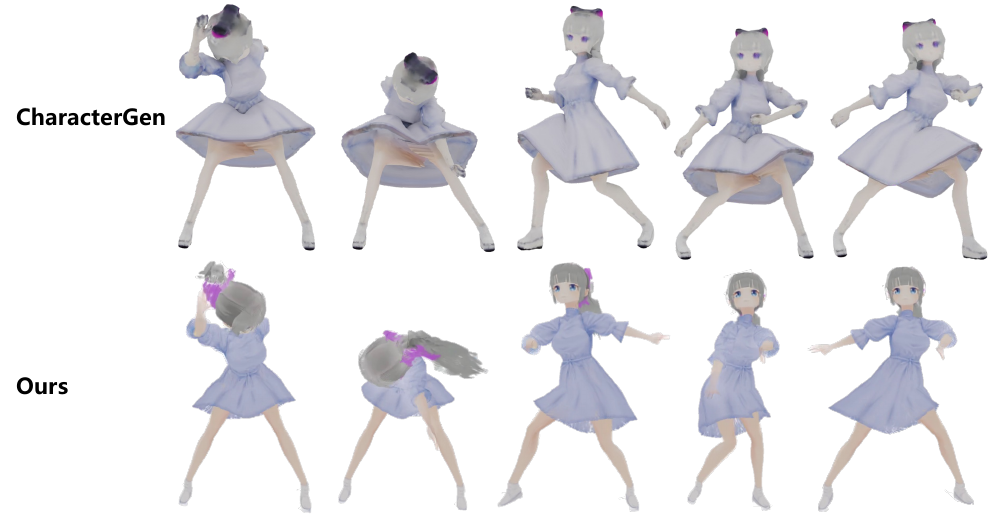}
  \caption{
We compare our method with CharacterGen~\cite{peng2024charactergen} animation results. 
Since \textbf{DRiVE} explicitly models clothing and hair, it generates more natural and realistic animations. 
Additionally, 3D Gaussian-based rendering achieves higher quality than mesh.
}
  \label{fig:demo}
\end{figure}

Nevertheless, the development of animating created 3D characters is relatively lagged -- the current main focus is still positioned on addressing motions of the human body, which is typically based on 2D~\cite{OpenPose} or 3D parametric human models~\cite{SMPL2015, SMPLX2019}. 
The lack of tailored-for designs for external parts beyond the human body significantly limits the downstream applications. 
For instance, as shown in Fig.~\ref{fig:demo}, naively animating a female character in a dress and wearing her hair in a ponytail with a body skeleton would lead to unrealistic rendering since her dress and hair are rigidly stuck to the body. 

In light of the above challenge, we present \textbf{DRiVE}, to the best of our knowledge, the first framework towards \emph{generation and animation} of 3D human characters with rich structures beyond body parts. 
We first discuss the critical design choices in building DRiVE as follows. 

\noindent\textbf{Animation Representation: }
While it is tempting to shoot for an automatic framework in a data-driven way, we take a step back and resort to computer graphics for a balance between automation and control degree. 
More specifically, we aim for \emph{rigging} the generated characters with a set of skeletons including rich structures such as hair and clothing and the corresponding skinning.
It is worth noting that SMPL and the follow-ups are essential rigging methods as well. 
The key difference in-between is that our rigging target is of a heterogenous structure, which is far more challenging than the former.


\noindent\textbf{Generation representation: }There is an amount of 3D shape representations of choice for generative models.
As we desire efficient animation and rendering, we rule out point clouds and implicit representations despite their advantage in representing the complex topology of garments. 
It may then seem natural to settle down at polygonal mesh, due to its popularity in the prior academic and industrial efforts on rigging. 
However, we identify 3D Gaussian~\cite{3dgs} as a better alternative for the following reasons: 1) making use of mesh-based rigging frameworks requires meshing quality (\emph{e.g.,} watertight~\cite{baran2007automatic}), which is hard to guarantee in generation; 2) even geometry of mesh were given a prior, producing high-quality texture maps remains an open problem~\cite{zeng2024paint3d} (also see in Fig.~\ref{fig:lgmfusion}). 

To conclude, DRiVE generates high-quality 3D Gaussian from low-level input such as a single image or text prompt. 
Then it assigns rigging information automatically to the 3D Gaussians, which can be further used for fine-grained control over the body but also external structures. 

Two core challenges surface in our design. 
First, to our knowledge, there does not exist any large-scale dataset for rigging characters with rich structures in general, letting alone being specific to 3D Gaussian; 
Second, apart from the essential difficulty of learning heterogeneous skeletons from geometries, our generated 3D Gaussian lacks geometric structure, which has been considered important in estimating skeletons~\cite{xu2019predicting, xu2020rignet}. 
In response to \textbf{Data Insufficiency}, we curate $\dn$, a large set of $9420$ meshes with calibrated rigging annotations (including skeleton and skinning). 
Then, to adapt to our representation, we fine-tune LGM~\cite{tang2024lgm} to generate 3D Gaussian from a single image. 
Finally, we perform label transfer from the original meshes to the corresponding 3D Gaussian. 
We remark that the fine-tuned LGM also serves as our final generation model; 
Regarding \textbf{Difficulty of Learning Skeletons on 3D Gaussian}, we propose a novel diffusion module, GSDiff, to de-noise for \emph{joint positions} conditioned on the input 3D Gaussian. 
In contrast to the previous regression-based methods, GSDiff treats joint positions as spatial distributions, taming the learning difficulty. 
Additionally, for fully exploiting 3D Gaussian, we separately extract geometric and appearance features from the means and canonical multi-view renderings for diffusion. 
The accurate joint prediction then lays a good basis for further estimating bone connection and skinning weight, leading to high-quality rigging results. 

Last but not least, beyond our best expectations, the above framework can be easily extended to rigging (potentially textured) meshes, which on its own is a tough task. 
More specifically, we \emph{drop} the edge connection of the meshes, and train GSDiff using the vertex sets and canonical multi-view renderings as we train on 3D Gaussian.

We summarize our main contributions as follows:
\begin{enumerate}
	\item We introduce a pipeline that generates rigged 3D models from multimodal inputs, enabling the detailed modeling of complex elements such as hair and clothing, thereby producing high-quality, free-viewpoint rendered videos.
	\item  We propose GSDiff, a novel 3D Gaussian-based diffusion network, to accurately predict joint positions. For 3D Gaussian conditional inputs, we specifically design a tailored conditioning approach.
	\item By fully exploiting 3D Gaussian geometric and appearance information, we enhance the rigging process, enabling precise skeleton binding and skinning estimation.
	\item We introduce $\dn$, a large-scale character rigging dataset, and achieve state-of-the-art results in the skeleton and skinning predictions on this dataset. Our results can be directly integrated into animation pipelines, significantly reducing the complexity of animators' workflows.
\end{enumerate}

\section{Related Works}
\subsection{3D Avatar Rigging}

In 3D avatar rigging, the task traditionally requires manual rigging by designers, which is both time-consuming and tedious. 
In recent years, with the advancement of machine learning technologies, automatic rigging methods have begun to emerge. 
Notably, Neural body~\cite{peng2021neural} first proposes using the SMPL to generate dynamic 3d human models automatically. 
This groundbreaking research laid the foundation for animatable 3d human generation using the SMPL model for~\cite{hu2023sherf, su2023caphy, xu2024xagen, cao2024dreamavatar}. 
However, SMPL model focuses entirely on the human body, which cannot effectively drive clothing, hair, and other elements, resulting in unrealistic dynamic effects. 
To this end, ~\cite{qin2020pointskelcnn} and ~\cite{yang2021s3} do not directly use SMPL but instead learn from rigging data with labeled human bodies. 
However, their labeled skeleton data still has a similar topology to SMPL, leading to similar issues during animation as seen in SMPL-based methods.

To address the aforementioned issues, we focus on generating heterogeneous skeletons. 
Current mainstream methods can be divided into two categories: optimization-based methods and learning-based methods. 
CASA~\cite{wu2022casa} is the first to propose jointly inferring articulated skeletal shapes and rigging through optimization. 
Subsequent follow-ups combine techniques such as dynamic NeRF~\cite{yang2022banmo, yang2023reconstructing, tan2023distilling} and dual-phase optimization~\cite{zhang2024s3o} to improve the quality of 3D object reconstruction and rigging. 
However, optimization-based methods can only be applied on a per-case basis, lacking generalization capabilities and therefore are costly for large-scale data processing.

Recent works have utilized learning-based methods to generate heterogeneous skeletons~\cite{baran2007automatic, xu2019predicting, xu2020rignet, yang2022object, ma2023tarig} and skinning~\cite{dionne2013geodesic, liu2019neuroskinning}. 
Among them, RigNet~\cite{xu2020rignet} takes a mesh as input and designs networks to predict the positions of joints, bone connections, and skinning separately. 
For joint estimation, RigNet first predicts offsets using a regression-based approach and then performs clustering in a differentiable manner, determining joint positions based on the cluster centers. 
This sequential pipeline is relatively complex and prone to error accumulation. 
Moreover, predictions based on regression methods tend to lack generalization capability. 
We propose a novel pipeline for joint prediction based on conditional diffusion. 
By learning the distribution of joints through diffusion, our method enables the generation of complex heterogeneous skeletons.

\subsection{3D Avatar Generation}
The development of general 3D generation technologies has significantly enhanced the realism and detail of avatars created from various inputs, ranging from objects to human figures~\cite{hong2023lrm, xu2024instantmesh, tochilkin2024triposr, boss2024sf3d, wu2024unique3d, long2024wonder3d, li2024craftsman, liu2024syncdreamer, li2024era3d, shi2023zero123plusplus, liu2024one, liu2023zero, zhang2024clay, chen2024idea}. 
However, human-centered 3D reconstruction methods focus on high-fidelity digitization and reconstruction of clothed humans from minimal inputs~\cite{saito2019pifu, dong2019fast, saito2020pifuhd, huang2024tech, xiu2022icon, xiu2023econ, zhang2024sifu, ho2024sith, xiu2024puzzleavatar, jiang2022selfrecon, peng2021animatable, hu2023sherf, chen2024ultraman}. 
Recent approaches have taken single-image 3D reconstruction of anime characters to new heights using cartoon datasets~\cite{peng2024charactergen}. 
The Gaussian-based approach offers better rendering quality compared to the mesh-based approach~\cite{shao2024splattingavatar, xu2023gaussian, liu2023humangaussian, hu2023gaussianavatar}. 
Despite these advancements, existing methods often overlook skeletal integration and typically use SMPL (Skinned Multi-Person Linear Model)~\cite{SMPL2015, SMPLX2019} as the driving skeleton, which can cause unnatural behavior in elements like skirts. 
To address this issue, we propose a method incorporating heterogeneous skeletons into Gaussian representations, allowing for a more realistic simulation of clothing and hair movements in sync with the avatar's body, which is crucial for gaming applications.

\section{Dataset Construction}\label{sec:data}
\begin{figure}[t!]
  \centering
  \includegraphics[width=\linewidth]{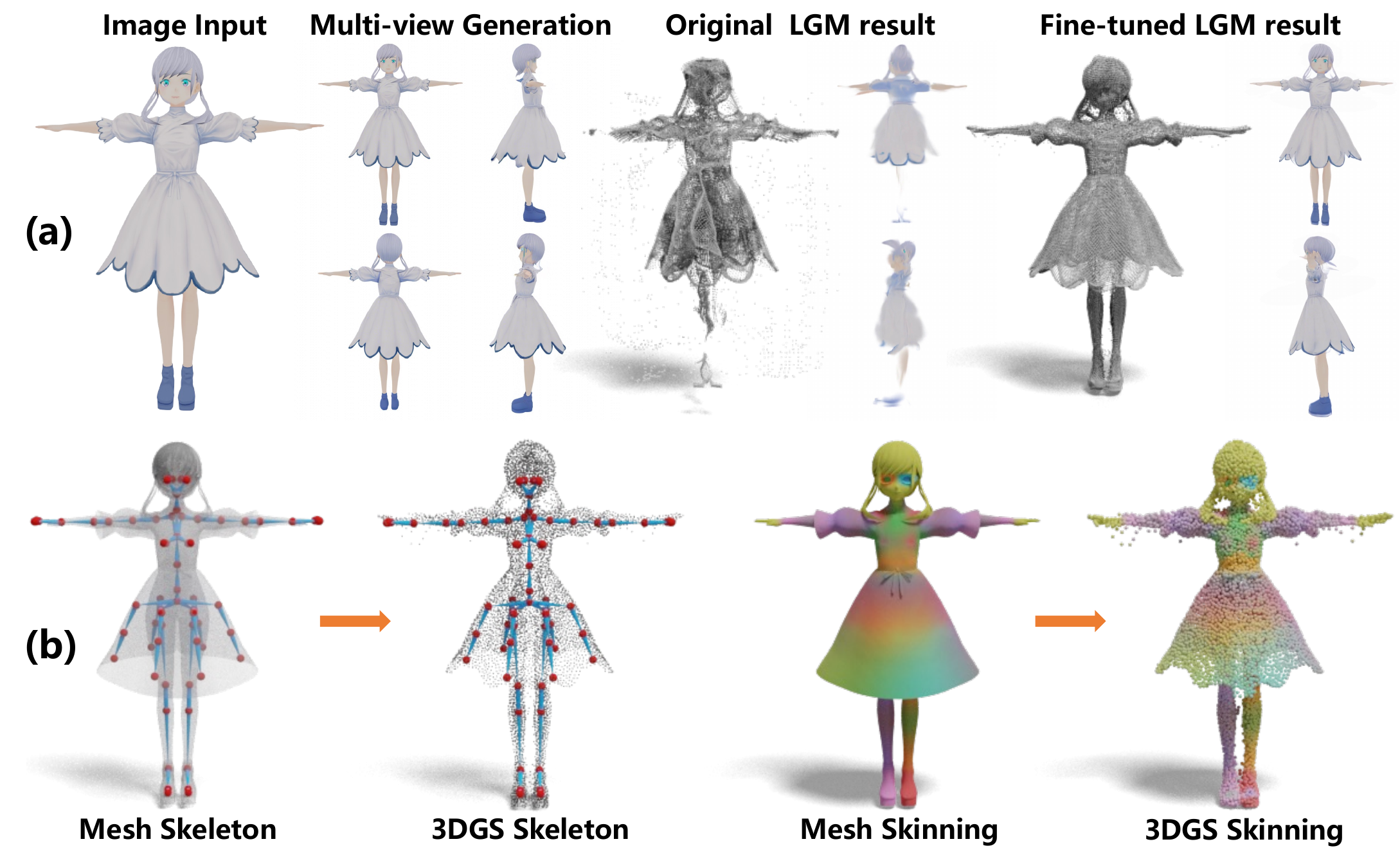}
    \caption{
    We show the comparison results of LGM before and after fine-tuning on our dataset in (a). We present the Ground Truth skeleton and skinning results of the mesh and the results transferred to 3D Gaussian in (b).}
  \label{fig:dataset}
\end{figure}

We construct our $\dn$ dataset with the following main stages and defer the technical details to Supp. Mat.: 
\begin{enumerate}
	\item \textbf{Data collection:} We first curate a large set of $13746$ textured meshes with initial rigging annotations from VRoidHub~\footnote{https://hub.vroid.com/en}, then we filter out the non-humanoid ones and ask artists to manually repair the significant errors, resulting in a subset of $9420$ meshes with reliable annotations. Moreover, each joint is accompanied by some semantic label, such as "J\_Bip\_L\_Hand, J\_Sec\_R\_SkirtBack1". 
	\item \textbf{Conversion to 3D Gaussian:} Then we convert the rigged meshes to 3D Gaussian representation. Specifically, we resort to fine-tuning LGM~\cite{tang2024lgm} with images rendered from the given meshes, which leads to a conversion route as input mesh $\rightarrow$ 4-view image rendering of the mesh $\rightarrow$ 3D Gaussian generated by fine-tuned LGM. 
	\item \textbf{Label Transfer to 3D Gaussian:} Since the faithfully generated 3D Gaussian do not necessarily align with the mesh input, we further employ a scaled Iterative Closest Point (ICP) algorithm~\cite{chetverikov2002trimmed} to register the mesh to the corresponding 3D Gaussian, with naturally transform annotations from the source meshes to the 3D Gaussian. An example is shown in Fig.~\ref{fig:dataset}(b).
\end{enumerate}

\section{Methodology}\label{sec:Methodology}

\begin{figure*}[t!]
  \centering
   \includegraphics[width=\linewidth]{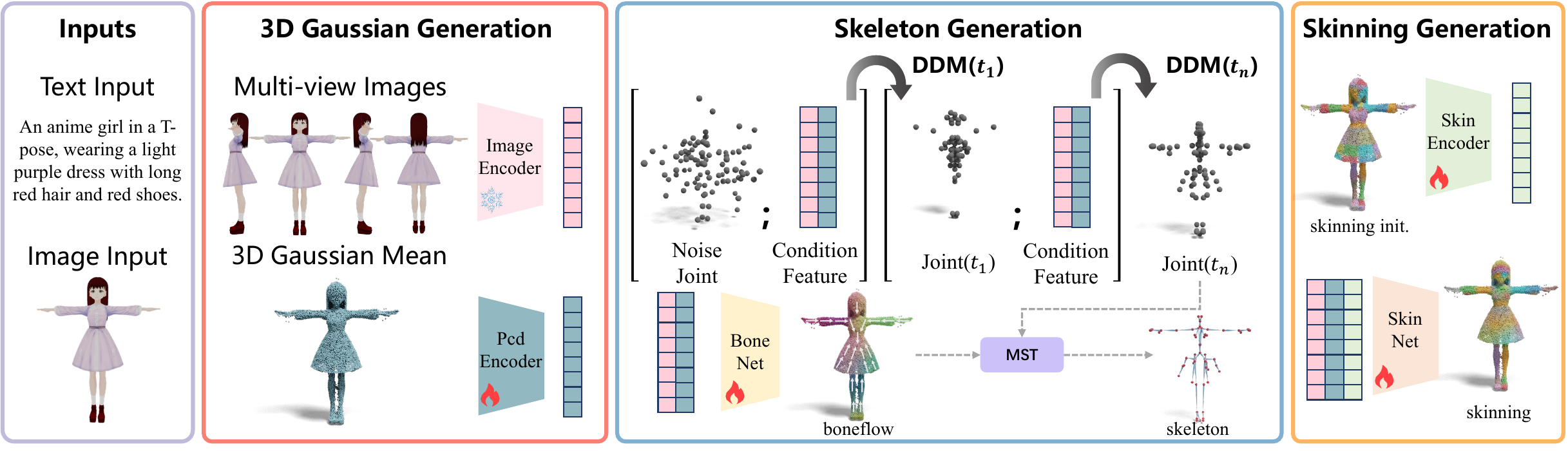}
  \caption{
  The overall pipeline of our framework. 
  See the main text for more details. 
  }
  \label{fig:overallpipeline}
\end{figure*}

In this section, we demonstrate the overall pipeline of our rigging system, which, trained on our $\dn$ dataset, can automatically and efficiently produce high-quality rigging results on 3D human characters with rich structures (\emph{e.g., clothing, hair}) originating from multi-modal inputs. 
In the following, we describe how we preprocess different inputs in Sec.~\ref{sec:meh_preprocess}, which are uniformly converted to a canonical input consisting of 3D Gaussian points and a set of images rendered from 4 views. 
Then in Sec.~\ref{sec:meh_rigging}, we sequentially go through our modules for predicting joints, connecting bones, and assigning skinning weights. 
Finally, tailored for the image inputs, in Sec.~\ref{sec:meh_Fine_Detail_Enhancement} we propose a simple yet effective module to boost the rendering quality of the rigged characters



\subsection{Input Pre-processing}\label{sec:meh_preprocess}
Our framework is flexible in accommodating various input modalities, including text prompts, and anime images. 
This is in contrast to the prior works~\cite{xu2019predicting, xu2020rignet}, which in general are limited to T-pose meshes. 
In the following, we briefly describe the pre-processing procedure tailored for each modality and refer the readers to Supp. Mat. for more details. 
As shown in Sec.~\ref{sec:data}, we have fine-tuned the LGM model on our $\dn$ dataset, which allows for generating faithful 3D Gaussian with a single image of a T-pose character from the frontal view. 
In the following, we describe how we obtain the desired image input given the following non-3D inputs: 

\noindent\textbf{Anime Image: }
We convert non-T-pose anime images to T-pose using our fine-tuned Animagine-XL~\footnote{https://huggingface.co/cagliostrolab/animagine-xl-3.1} model with IP-Adapter~\cite{ye2023ip} and OpenPose~\cite{OpenPose} ControlNet~\cite{zhang2023adding}. This enables standardized T-pose anime image input, allowing the LGM model to generate accurate 3D Gaussian.


\noindent\textbf{Text Prompt: }We fine-tune Animagine-XL model with T-pose images obtained from our $\dn$ dataset, which can generate a frontal image of a T-pose character that matches the input text description. 


After generating a 3D Gaussian, we extract 3D Gaussian points from the means and render images from four canonical views: front, back, left, and right.


\subsection{Rigging for Gaussian Character}\label{sec:meh_rigging}

Our rigging pipeline is divided into two parts as shown in Fig.~\ref{fig:overallpipeline}.  
In Sec.~\ref{sec:meh_rigging_joints}, we introduce how to predict the positions of joints using a diffusion model based on both geometric and visual cues.
We also determine the connections between unordered joints through Minimum Spanning Tre (MST). 
Finally, we estimate the corresponding skinning with respect to the predicted joint positions in Sec.~\ref{sec:meh_rigging_skinning}.

In particular, We train the pipeline with the rigged 3D Gaussian from our proposed $\dn$ dataset. 
Given a 3D Gaussian, hereafter we denote by $\mu$ the means, and by $\{I_i\}^4_{i = 1}$ the images rendered from 4 canonical views.

\subsubsection{Skeleton Generation}\label{sec:meh_rigging_joints}
A key challenge in rigging 3D human characters with varying hair and clothing styles arises from the diversity in joint distributions.
For instance, a female character in an intricate dress requires more joints to drive than a male character in swimming trunks. 
The skeletal heterogeneity prevents one from learning a direct mapping from input geometry to joints. 
Prior works~\cite{xu2020rignet, ma2023tarig} then take an indirect approach, which learns per-point features on input geometry with graph neural networks, and then leverages a non-learnable clustering step to contract surface points towards the labeled joints with regression loss as guidance.  
On the other hand, since garments are often spatially close to the body, mesh connectivity (\emph{i.e., }topology) is of critical importance in disentangling their features. 
The above methods therefore depend on high-quality mesh, which is hard to guarantee in our task of interest. 

In response to the lack of topology in 3D Gaussian representation, we propose a novel module, \textbf{GSDiff}, a 3D \textbf{G}aussian \textbf{S}platting conditioned \textbf{Diff}usion model for estimating heterogeneous skeletons on our generated 3D Gaussians.  
To fully exploit 3D Gaussian, \textbf{GSDiff} takes the means $\mu$ and multi-view rendering $\{I_i\}_{i = 1}^4$ of a 3D Gaussian as input, which carry respectively the geometric and visual information. 
More specifically, we aim to learn to denoise for the joints, $\mathbf{J}$, conditioned on the above input.


We start by introducing a plain diffusion model to learn the joint distribution conditioned on 3D Gaussian input $q(\mathbf{J} |\mu,\{I_i\}^4_{i = 1})$. 
We can sample from $q(\mathbf{J} |\mu,\{I_i\}^4_{i = 1})$ by starting with noise input and then iteratively sampling from $q(\mathbf{J}_{t-1}|\mathbf{J}_{t},\mu,\{I_i\}^4_{i = 1}))$.
During the reverse process, we use a network $s_\theta$ to approximate the conditional distribution:
\begin{equation}\label{eqn:1}
\begin{aligned}
& s_\theta(\mathbf{J}_{t-1}|\mathbf{J}_{t},\mu,\{I_i\}^4_{i = 1}) \approx q(\mathbf{J}_{t-1}|\mathbf{J}_{t},\mu,\{I_i\}^4_{i = 1}).
\end{aligned}
\end{equation}
We train a DGCNN~\cite{phan2018dgcnn} as $\mathbf{DG}$ and obtain features for $\mu$.  
We define the $i^{th}$ joint in the step $t$ as $\mathbf{J}_{t}^{(i)} \in \mathbb{R}^{1 \times 3}$.

Though conditional diffusion has been widely used in various tasks, the denoising object and the conditional input are often comparable (either of the same modality~\cite{latentdiffusion} or can be associated by simple operations like rendering~\cite{tyszkiewicz2023gecco, melas2023pc2}).  
Our task, on the other hand, falls into a new category. 
Although both $\mu$ and $\mathbf{J}$ are point clouds, the latter is effectively an \emph{abstract} of the former, the associating operation is exactly what we seek at the beginning. 

We address this problem by proposing a novel conditional diffusion scheme, which enhances the interaction between the conditional inputs and the intermediate denoising outcome.  
More concretely, considering $\mathbf{J}_{t}^{(i)}$, the $i-$th joint at the $t-$th step of denoising process. 
We search for the $k-$nearest neighbors of it among $\mu$, and denote by $\{ id_{t,i}^{(1)}, \ldots, id_{t,i}^{(k)} \}$ the regarding indices. 
Then we compute the geometric feature of $\mathbf{J}_{t}^{(i)}$, $F(\mathbf{J}_{t}^{(i)})$ as follows:
\begin{equation}\label{eqn:2}
F(\mathbf{J}_{t}^{(i)}) = \frac{\sum_{j=1}^{k} w_j \mathbf{DG}(\mu_{id_{t,i}}^{(j)})}{\sum_{j=1}^{k} w_j},
\end{equation}
where $id_{t,i}^{(l)}$ represents the index of the $l^{th}$ nearest neighbor, $w_j = \frac{1}{\| \mathbf{J}_t^{(i)} - \mu_{id_{t,i}}^{(j)} \|}$ and $\| . \|$ denotes the Euclidean distance. 
This way we can get $F(\mathbf{J}_{t}) \in \mathbb{R}^{m \times 128}$ for $m$ joints which integrates information from the surrounding 3D Gaussian, making it easier to determine the position of each joint within the 3D Gaussian accurately. 

Moreover, we use CLIP~\cite{clip} to get the appearance condition feature based on the multi-view images $\{I_i\}^4_{i = 1}$.
We combine the geometry and appearance feature with the joint feature as the input to each denoising step. 
We can sample the joint $\mathbf{J}$ by iteratively sample from $s_\theta(\mathbf{J}_{t-1}|F(\mathbf{J}_{t}), \mathbf{CLIP}(\{I_i\}^4_{i = 1}))$.
Additionally, inspired by~\cite{podellsdxl}, we add a cross-attention layer after each self-attention layer. 
In particular, we use a modified set transformer as the backbone of the diffusion model~\cite{tyszkiewicz2023gecco}.

Last but not least, thanks to the rich semantic labels in $\dn$, we separate a set of $25$ joints corresponding to body parts from each character. 
Therefore, $\textbf{J}$ is further divided into two parts: body joints $\mathbf{J}_{b}$, and otherwise $\mathbf{J}_{o}$, which includes hair, clothing, and other un-common parts specific to some character. 
For ease of learning, we train two separated diffusion models to predict $\mathbf{J}_{b}$ and $\mathbf{J}_{o}$ respectively using the same condition.

\noindent\textbf{Bone Connection: }Based on our denoised joints, we follow TARig~\cite{ma2023tarig} to train a BoneFlow for assisting bone connection. 
To construct BoneFlow on ground-truth joints, we find for each surface point $p$ its nearest neighbor in joints and the regarding parent.  
Then BoneFlow at $p$ is defined as the vector pointing from its nearest neighbor to its parent joint. 
Similarly, we set $\mu$ and $\{I_i\}_{i=1}^4$ as input to a network learning to predict BoneFlow. 
Then we can perform MST algorithm~\cite{prim1957shortest} on a cost matrix built on BoneFlow and our estimated joints to obtain bone connection. 
We refer the reader to the Supp. Mat. for the implementation details.

\subsubsection{Skinning Generation}\label{sec:meh_rigging_skinning}
The final step is to estimate the skinning based on the predicted skeleton. 
Prior arts depend on geometric cues (\emph{e.g., }geodesic distances~\cite{xu2020rignet}) for skinning. 
Unfortunately, such is unavailable for 3D Gaussian. 

However, thanks to our accurate estimation of the skeleton in the last section, we empirically observe that performing \text{k-NN} searches between the joints and the 3D Gaussian and constructing $\mathbf{S}_{init}\in \mathbb{R}^{n\times m}$ as the distance-based similarity matrix already yields a decent estimation of skinning. 
Here $n, m$ are respectively the number of 3D Gaussian points and joints. 
We defer the details to the Supp. Mat. 
We then use $\mathbf{S}_{init}$ as the initial estimate for skinning and combine it with 3D Gaussian features to predict the ground truth skinning.
\begin{equation}\label{eqn:4}
\hat{\mathbf{S}} = f_{s}(\mu,\{I_i\}^4_{i = 1},\mathbf{S}_{init};\mathbf{W}_s),
\end{equation}
where $\mathbf{W}_s$ denotes the learned parameters of the skinning estimation. 

By treating the per-vertex skinning weights as probability distributions, we use cross-entropy ($\mathcal{L}_{ce}$) and Kullback-Leibler divergence ($\mathcal{L}_{kl}$) as the loss function to measure the disagreement between the ground truth and predicted distributions for each vertex. 
Since we aim to animate the dense 3D Gaussian, the generated skinning needs to change smoothly to prevent inconsistencies in skinning between adjacent joints, which could cause cracks during the animation.
So we introduce a smooth loss as a regularization term.
\begin{equation}\label{eqn:5}
\mathcal{L}_{smooth} = \frac{1}{n} \sum_{i=1}^{n} \frac1{|\mathbb{D}_i|}\sum_{j\in\mathbb{D}_i} \left\| \hat{\mathbf{S}}_{i} - \hat{\mathbf{S}}_{j} \right\|_2,
\end{equation}
\begin{equation}\label{eqn:6}
\mathcal{L}_{skinning}=\mathcal{L}_{ce}+\mathcal{L}_{kl}+\lambda_{2}\mathcal{L}_{smooth},
\end{equation}
where $\hat{\mathbf{S}}_{i}$ denotes the predicted skinning of vertex $i$, $\mathbb{D}_i$ represents the set of Gaussian points surrounding vertex $i$ and $n$ is the number of 3D Gaussian vertices.

\subsection{3D Gaussian Refinement}\label{sec:meh_Fine_Detail_Enhancement}

\begin{figure}[htbp]
  \centering
  \includegraphics[width=\linewidth]{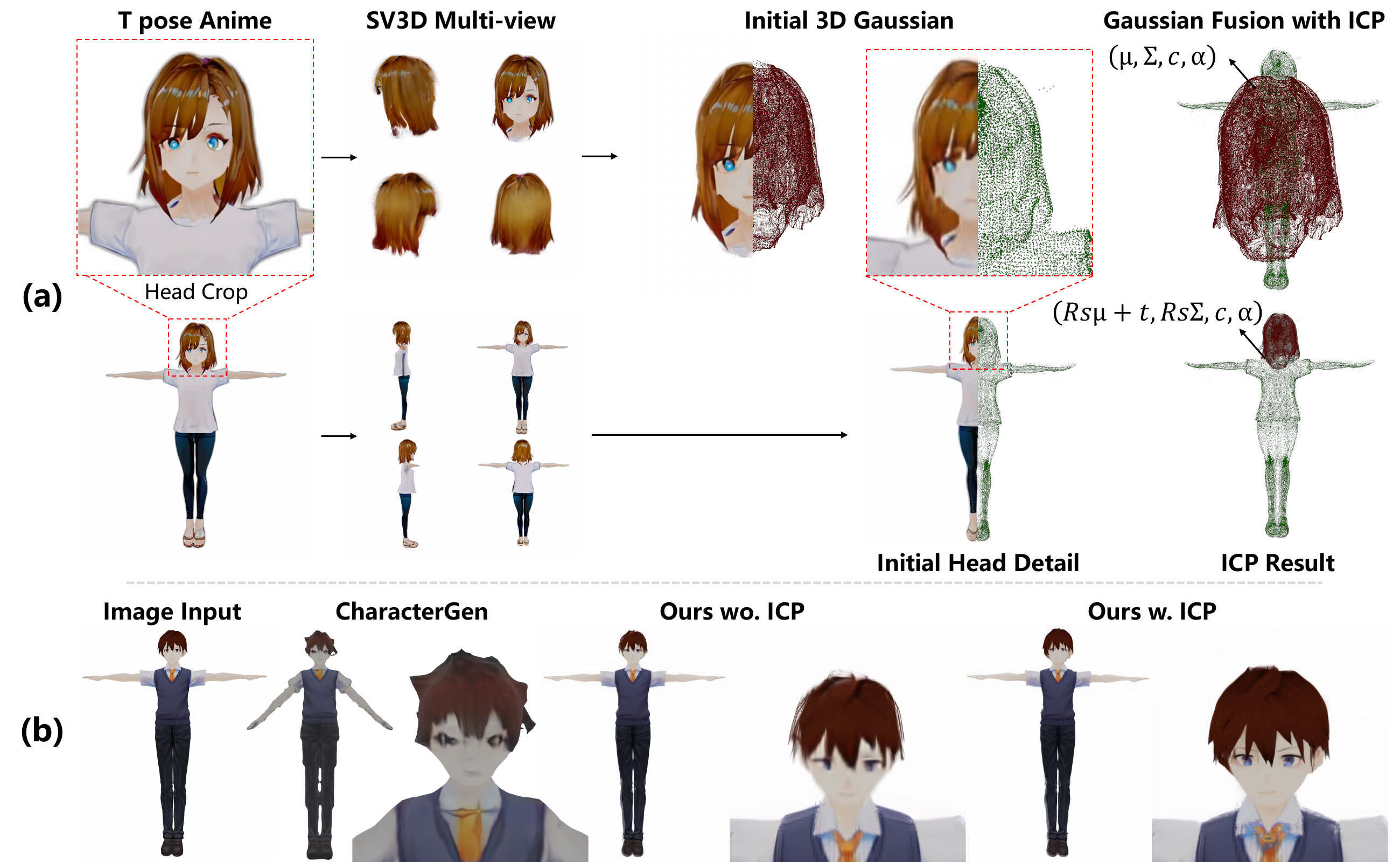}
  \caption{Pipeline for 3D Gaussian refinement and results using a T-pose anime image.}
  \label{fig:lgmfusion}
\end{figure}

In this step, we use a T-pose anime image as a condition to generate 3D Gaussian points. 
To address the severe artifacts and loss of detail in the head region observed in the original LGM results (Fig.~\ref{fig:lgmfusion}(a)) when using a full-body image as input, we fine-tune LGM on our $\dn$ dataset and perform separate reconstructions for the head and body. 
Using SV3D~\cite{voleti2024sv3d} to enhance consistency, we generate new view images at 15° intervals horizontally and select four images at 90° intervals, including the frontal view, as input for LGM. 
The 3D Gaussian points for the head and body are initially aligned using a fixed cropping frame, refined with the Iterative Closest Point (ICP) algorithm~\cite{chetverikov2002trimmed}, and merged into a single representation. 
Overlapping points between the head and body Gaussians are filtered out to ensure a smooth transition.
Fig.~\ref{fig:lgmfusion}(b) shows the results before and after ICP refinement, alongside a comparison with CharacterGen, highlighting the improved alignment and detail preservation achieved by our method. 
Further details about the refinement stage are provided in the Supp. Mat.

\section{Experimental Results}

\subsection{Metircs and Baselines}
To evaluate the predicted skeletons and skinning maps, we adopt the same metrics as RigNet~\cite{xu2020rignet}.
For skeleton evaluation, we use CD-J2J (The Chamfer distance between joints), CD-J2B (The Chamfer distance between joints and bones), CD-B2B (The Chamfer distance between bones), IoU (Intersection over Union), and Precision \& Recall. 
For skinning evaluation, we leverage Precision \& Recall and L1-norm between predicted and reference skinning weights. 
For more details, please refer to the Supp. Mat.


\noindent\textbf{Skeleton prediction: }We compare our method with (1) AnimSkelVolNet~\cite{xu2019predicting} w. Ground Truth Mesh; (2) RigNet~\cite{xu2020rignet} w. Ground Truth Mesh; (3) RigNet w. Mesh from CharacterGen~\cite{peng2024charactergen}.
Specifically, the first two baselines use Ground Truth mesh as input to predict skeleton results with AnimSkelVolNet and RigNet. 
For a fair comparison, we also apply CharacterGen~\cite{peng2024charactergen} to produce high-quality 3D mesh predictions from a single character image, which we then use with RigNet for skeleton prediction (the third baseline).
Note that AnimSkelVolNet and RigNet are trained on the same dataset as our method.


\noindent\textbf{Skinning prediction: }We compare our method with (1) GeoVoxel~\cite{dionne2013geodesic} w. Ground Truth Mesh; (2) RigNet w. Ground Truth Mesh. 
For GeoVoxel, we utilize the implementation provided by Maya~\cite{Autodesk}. 
For all methods, we follow RigNet by using the Ground Truth skeleton as input to predict skinning during training and testing.



\subsection{Evaluation for Skeleton and Skinning}
\begin{figure}[htbp]
  \centering
  \includegraphics[width=\linewidth]{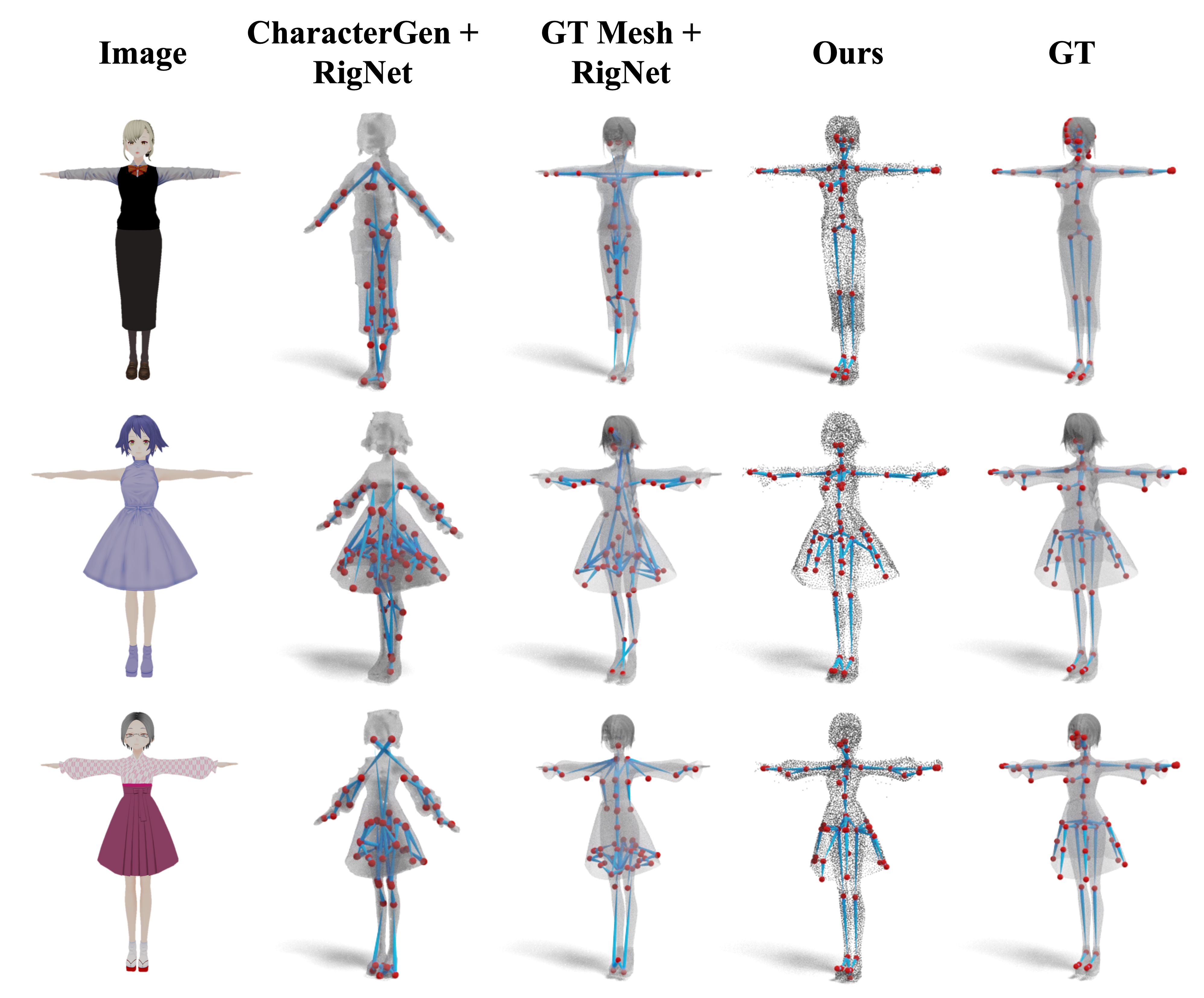}
  \caption{Our method accurately predicts skeletal structures, outperforming RigNet~\cite{xu2020rignet} in joint and bone estimation.}
  \label{fig:joints}
\end{figure}

\noindent\textbf{Skeleton evaluation:} Tab.~\ref{table:joints} presents the quantitative comparison in joint estimation. 
Our results significantly outperform AnimSkelVolNet and RigNet across all metrics. 
Specifically, the IoU metric, which measures the quality of joint estimation, shows a 59.3\% improvement, while the CD-B2B metric, which evaluates the accuracy of bone estimation, shows a 39.9\% improvement. 
For qualitative evaluation, we select examples featuring a variety of genders, hairstyles, and clothing styles, as shown in Fig.~\ref{fig:joints}. 
RigNet encounters difficulties in accurately estimating skeleton positions, particularly for clothing and hair regions. 
When using meshes predicted from a single image, challenges such as low mesh quality can even lead to joint estimation failures, such as predicting only one leg.
In contrast, our method generates significantly more plausible skeletons based on the predicted 3D Gaussians from a single image. 
While training RigNet on our dataset, we observe that both the vertex attention network and the final joint prediction network struggle to effectively converge with the more complex skeletal structure. 
This further highlights the advantages of our approach over regression-based methods.






\begin{table}[htbp]
\centering
\setlength{\tabcolsep}{4pt} 
\resizebox{\columnwidth}{!}{ 
\begin{tabular}{ccccccc}
\hline
\rowcolor[HTML]{FFFFFF} 
& IoU $\uparrow$             & Prec. $\uparrow$           & Rec. $\uparrow$            & CD-J2J $\downarrow$          & CD-J2B $\downarrow$          & CD-B2B $\downarrow$          \\ \hline

\rowcolor[HTML]{FFFFFF} 
AnimSkelVolNet      & 27.74\%         & 29.86\%         & 28.34\%     & 6.45\%           & 4.55\%           & 3.87\%                \\
\rowcolor[HTML]{FFFFFF} 
RigNet      & 28.69\%         & 23.81\%         & 38.13\%     & 5.37\%           & 3.72\%           & 3.26\%                \\
\rowcolor[HTML]{E8E8E8} 
Ours                         & \textbf{70.48\%} & \textbf{70.29\%} & \textbf{71.91\%} & \textbf{2.81\%} & \textbf{2.17\%} & \textbf{1.96\%} \\ \hline
\end{tabular}
}
\caption{
Joint prediction results on the test set.
AnimSkelVolNet and RigNet are results based on the Ground Truth Mesh.
}\label{table:joints}
\end{table}


\vspace{-4mm}

\begin{figure}[h!]
  \centering
  \includegraphics[width=\linewidth]{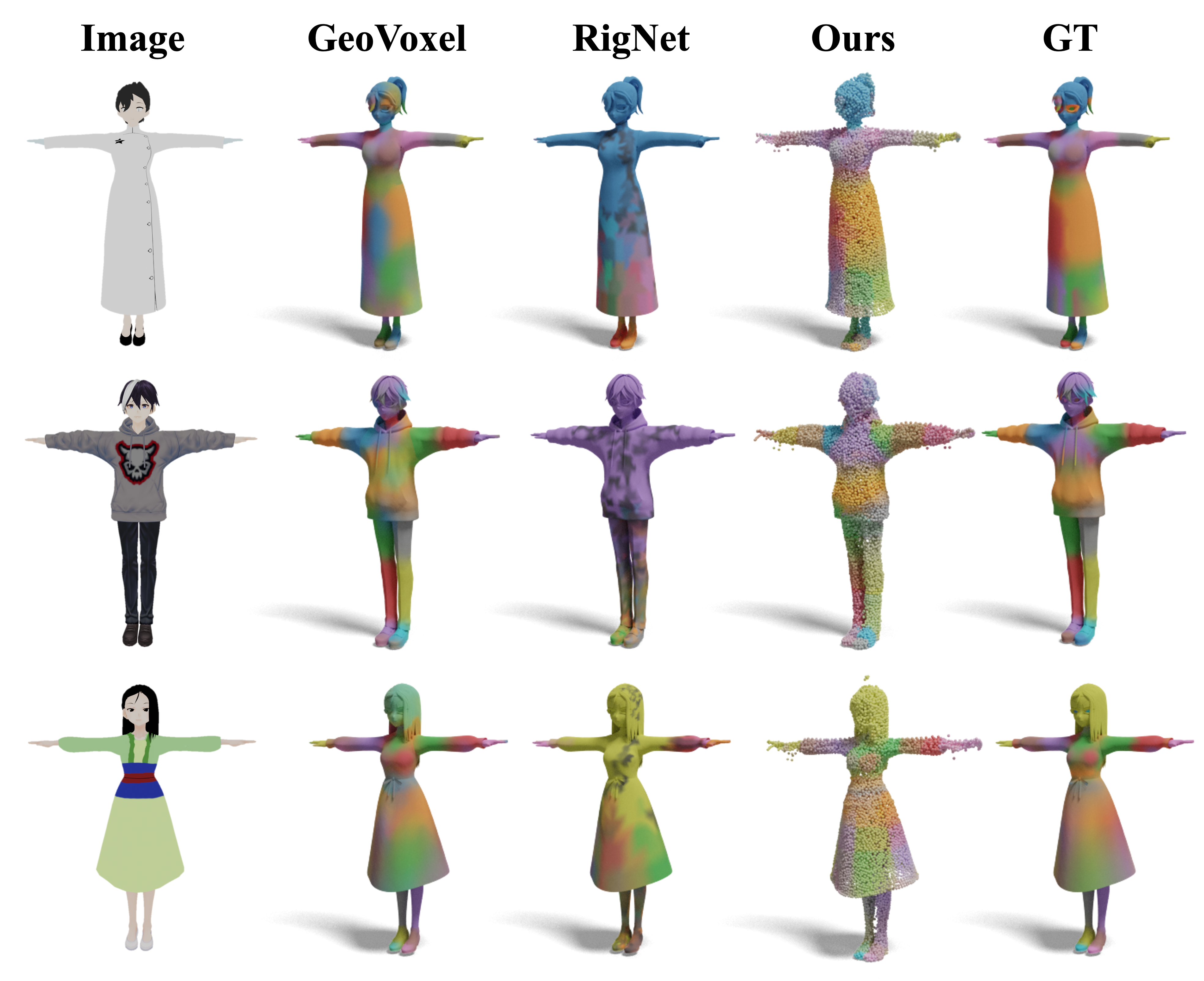}
  \caption{Our skinning predictions closely match the ground truth, surpassing Geovoxel~\cite{dionne2013geodesic} and RigNet~\cite{xu2020rignet}.}
  \label{fig:skinning}
\end{figure}

\noindent\textbf{Skinning evaluation: } Tab.~\ref{table:skinning} presents the evaluation metrics for skinning. 
Across all metrics, our results significantly outperform GeoVoxel and RigNet, with a 43.2\% improvement in precision and a 45.5\% reduction in average L1 error. 
Note that, due to the non-watertight nature of the mesh, the geodesic distance for RigNet cannot be calculated, so we use Euclidean distance instead. 
RigNet’s low recall in skinning prediction indicates its limitations in accurately predicting control points.
Fig.~\ref{fig:skinning} provides a qualitative comparison between our method and the baselines.
Our results align more closely with the ground truth overall. 
While GeoVoxel shows reasonable alignment with the Ground Truth in larger regions, it struggles to capture finer details, and RigNet often fails to produce reasonable results. 
These qualitative observations are consistent with the quantitative evaluation.


\begin{table}[htbp]
\centering
\setlength{\tabcolsep}{10pt} 
\resizebox{\columnwidth}{!}{ 
\begin{tabular}{cccc}
\hline
\rowcolor[HTML]{FFFFFF} 
                  & Prec. $\uparrow$           & Rec. $\uparrow$            & avg L1 $\downarrow$              \\ \hline
\rowcolor[HTML]{FFFFFF} 
GT Mesh w. GeoVoxel & 44.48\%          & 69.60\%          & 0.88                  \\
\rowcolor[HTML]{FFFFFF} 
GT Mesh w. RigNet & 41.94\%          & 35.89\%          & 1.00                 \\
\rowcolor[HTML]{E8E8E8} 
Ours              & \textbf{78.34\%} & \textbf{71.66\%} & \textbf{0.48}  \\ \hline
\end{tabular}
}
\caption{Skinning prediction results on the test set.}\label{table:skinning}
\end{table}

\subsection{Ablation study}

\noindent\textbf{Generation beats Regression: }
Here, we discuss the performance of joint position estimation using regression-based and generation-based approaches. 
As a baseline, we implement a regression-based method for estimating joint positions by replacing RigNet’s mesh encoder with the current state-of-the-art point cloud encoder~\cite{wu2022point} and creating a pipeline similar to RigNet for regressing joint positions from 3D Gaussians. 
We train and test this model on the same dataset, with the quantitative results for joint prediction presented in the first row of Tab.~\ref{table:joints_abl}.
Our observations show that the diffusion-based method significantly outperforms the regression-based approach in joint estimation. 
Regression methods face challenges in learning the positional information of skeletons with varying topologies, whereas diffusion-based methods, by learning positional distributions, exhibit a clear advantage.

\begin{table}[!t]
\centering
\setlength{\tabcolsep}{4pt} 
\resizebox{\columnwidth}{!}{ 
\begin{tabular}{ccccccc}
\hline
\rowcolor[HTML]{FFFFFF} 
                       & IoU $\uparrow$             & Prec. $\uparrow$           & Rec. $\uparrow$            & CD-J2J $\downarrow$          & CD-J2B $\downarrow$          & CD-B2B $\downarrow$          \\ \hline
\rowcolor[HTML]{FFFFFF} 
Regression      & 46.72\%         & 47.14\%         & 47.05\%     & 4.02\%           & 3.07\%           & 2.59\%                \\ 
\hline
\rowcolor[HTML]{FFFFFF} 
Ours w/o $C_{3d}$          &  56.56\%                &  56.18\%                &  58.12\%                  &  3.36\%               & 2.29\%                &  2.39\%               \\
Ours w/o $C_{3dl}$ &  58.71\%                &  59.18\%                &  59.49\%                 & 4.14\%                &  3.32\%               & 2.82\%                \\
\rowcolor[HTML]{FFFFFF} 
Ours w/o $C_i$          &  67.14\%                &  68.02\%                &    67.53\%              &  2.96\%               & 2.55\%                &  2.24\%               \\
\rowcolor[HTML]{E8E8E8} 
Ours                         & \textbf{70.48\%} & \textbf{70.29\%} & \textbf{71.91\%} & \textbf{2.81\%} & \textbf{2.17\%} & \textbf{1.96\%} \\ \hline
\end{tabular}
}
\caption{Ablation study on joint estimation. $C_{3d}$ denotes 3D Gaussian condition, $C_{3dl}$ denotes 3D Gaussian local condition, and $C_{i}$ denotes image condition.}\label{table:joints_abl}
\end{table}

\noindent\textbf{Conditional generation: }
We discuss the impact of different conditioning methods on the generated results during the diffusion-based joint generation process, with outcomes presented in Tab.~\ref{table:joints_abl}.
First, we conduct an ablation study without any 3D information (only image inputs), resulting in a 19.8\% drop in the IoU metric. 
This emphasizes the importance of incorporating geometric information from the 3D Gaussian in joint position estimation. 
We then test without the 3D Gaussian local condition that directly uses the 3D Gaussian point global feature as input, leading to a 16.7\% decrease in the IoU metric. 
This result suggests that local information from the 3D Gaussian is essential for accurately associating with joint data. 
Finally, when we remove the image input, we observe a slight decrease in performance, indicating that appearance information also contributes to joint estimation.
Overall, these ablation experiments validate the effectiveness of each component in our conditioning approach.

We experiment with replacing the input of \textbf{GSDiff} with mesh, using mesh as input for both training and testing.
As shown in Tab.~\ref{table:joints_mesh}, both inputs yield similar results, demonstrating the flexibility of our method.
\begin{table}[!t]
\centering
\setlength{\tabcolsep}{4pt} 
\resizebox{\columnwidth}{!}{ 
\begin{tabular}{ccccccc}
\hline
\rowcolor[HTML]{FFFFFF} 
                       & IoU ↑             & Prec. ↑           & Rec. ↑            & CD-J2J ↓          & CD-J2B ↓          & CD-B2B ↓          \\ \hline

\rowcolor[HTML]{FFFFFF} 
Plain Mesh      & 67.90\%         & 69.40\%         & 67.73\%     & 3.02\%           & 2.31\%           & 2.04\%                \\
\rowcolor[HTML]{FFFFFF} 
Textured Mesh      & 71.05\%         & 70.78\%         & 72.32\%     & 2.38\%           & 1.87\%           & 1.71\%                \\
\rowcolor[HTML]{FFFFFF} 
3D Gaussian                        & 70.48\% & 70.29\% & 71.91\% & 2.81\% & 2.17\% & 1.96\% \\ \hline
\end{tabular}}
\caption{Quantitative results of joints prediction using mesh and 3D Gaussian as input.}\label{table:joints_mesh}
\vspace{-4mm}
\end{table}
 

\noindent\textbf{Skinning generation: }
We also conduct experiments on the impact of different inputs for skinning estimation as shown in Tab.~\ref{table:skinning_abl}. 
First, we remove the rendered image input from 3D Gaussian, and the final average L1 error increased by 30.4\%, indicating that the appearance information from 3D Gaussian is crucial for skinning learning. 
We then experiment by removing the initial skinning values, which also led to an increase in average L1 error of 26.2\%, demonstrating that $S_{init}$ is essential for the skinning learning process. 
Finally, we show that the smoothness of 3D Gaussian skinning learning can also help the network converge.


\begin{table}[!ht]
\centering
\setlength{\tabcolsep}{10pt} 
\resizebox{\columnwidth}{!}{ 
\begin{tabular}{cccc}
\hline
\rowcolor[HTML]{FFFFFF} 
                  & Prec. $\uparrow$            & Rec. $\uparrow$            & avg L1 $\downarrow$           \\ \hline
\rowcolor[HTML]{FFFFFF} 
Regression & 39.46\%          & 33.93\%          & 1.32               \\
\hline
\rowcolor[HTML]{FFFFFF} 
Ours w/o image input & 62.93\%          & 71.01\%          & 0.69                  \\
Ours w/o $S_{init}$ input & 57.46\%          & 70.48\%          & 0.65                  \\
Ours w/o smooth loss & 77.36\%          & 71.31\%          & 0.49                 \\
\rowcolor[HTML]{E8E8E8} 
Ours              & \textbf{78.34\%} & \textbf{71.66\%} & \textbf{0.48}  \\ \hline
\end{tabular}}
\caption{Ablation study on skinning prediction. $S_{init}$ denotes initial skeleton input.}\label{table:skinning_abl}
\vspace{-4mm}
\end{table}

\subsection{Applications}\label{Applications}
\begin{figure}[h!]
  \centering
  \includegraphics[width=\linewidth]{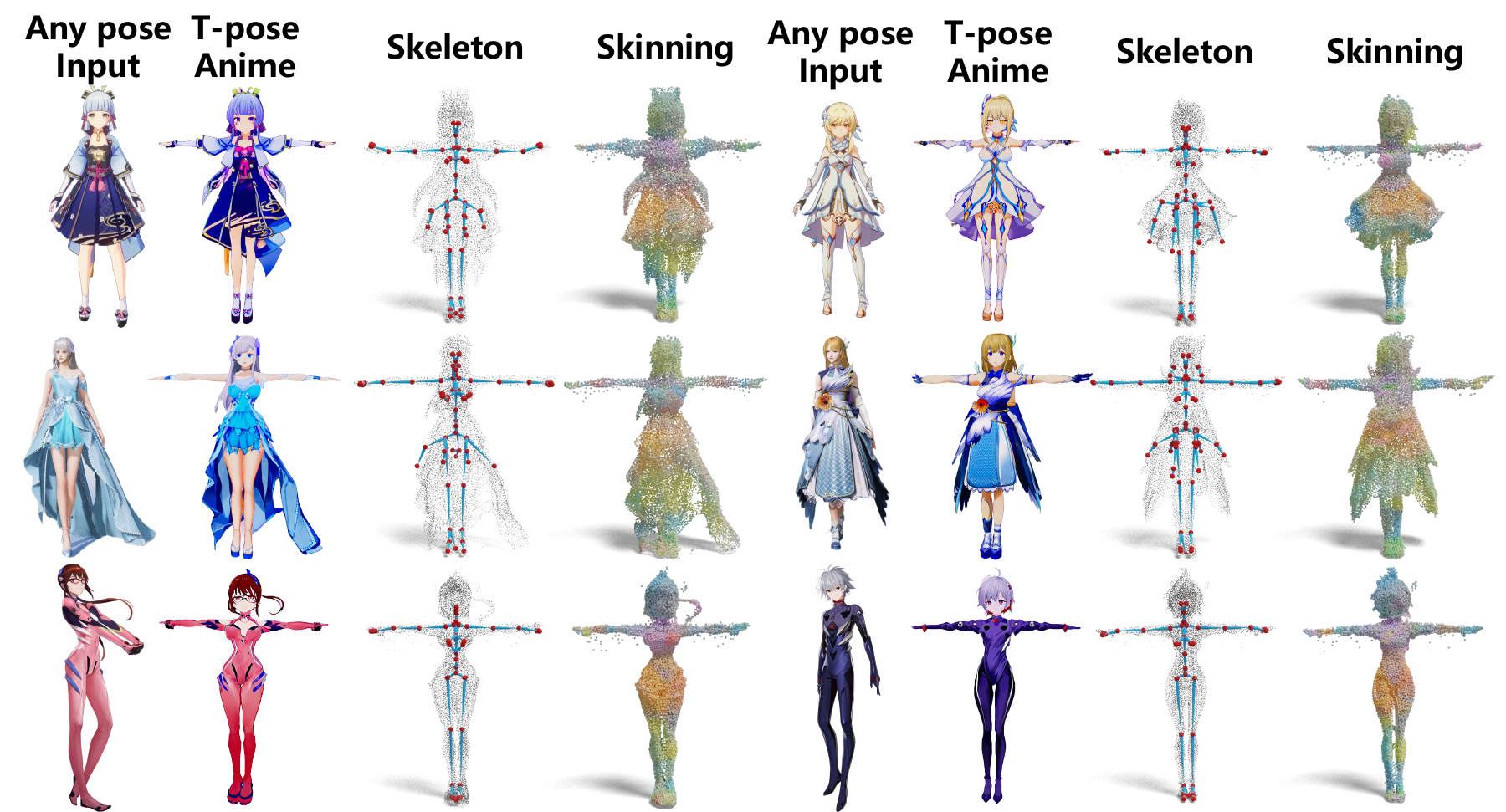}
  \caption{
  Our pipeline can consistently convert anime characters in any pose to T-pose and provide a reliable rigging result.
  }
  \label{fig:rigging_anypose}
\end{figure}

\noindent\textbf{Rigging with Any Pose Anime Image Inputs: }
In this section, we show an application of rigging from an arbitrary pose anime image to the final rigging results. 
We select image data from~\cite{peng2024charactergen}, which has a certain domain gap compared to our dataset. 
We first consistently convert the input image to a T-pose, then reconstruct the 3D Gaussian, and finally generate the corresponding rigging results, as shown in Fig.~\ref{fig:rigging_anypose}. 
Our method successfully produces reasonable skeletons and skinning for various clothing and hairstyles, demonstrating the generalization capability and applicability of our approach.

\noindent\textbf{Animation for rigged characters: }Here, we present the results of animating the rigged 3D Gaussian model as shown in Fig.~\ref{fig:demo}. Compared with CharacterGen, our 3D Gaussian-based method produces higher-resolution rendered results. Additionally, the generated skeletons for the skirt and hair display more natural movement, effectively preventing issues like the skirt and legs moving in unison.
\section{Conclusion, Limitation, and Future Work}
In this work, we introduce \textbf{DRiVE}, the first framework for generating and rigging 3D characters with rich structures using 3D Gaussian representations.
Our approach achieves state-of-the-art accuracy in skeleton and skinning predictions, enabling detailed modeling of hair and garments for realistic animations. 
By curating a large-scale rigged character dataset, $\dn$, and proposing the novel diffusion module \textbf{GSDiff}, we effectively address the challenges of data insufficiency and skeleton prediction on 3D Gaussian. 
The flexibility of our framework allows it to be extended beyond 3D Gaussian to mesh, further demonstrating its versatility. 
We believe DRiVE offers valuable insights for rigging and animation, with promising potential for practical applications.

We also identify the following limitations, which lead to future work directions: 
1) The node density of our generated skeleton is deterministic, and cannot be tuned according to user preferences. It would be interesting to learn from more versatile data; 
2) Unlike meshes, there is no mature solution for 3D Gaussian collision detection. Thus our rigged 3D Gaussian can suffer from self-crossing among parts (\emph{e.g., }leg and dress). 
While the high rendering frame rate allows for efficient human inspection, it would be interesting to explore more automatic solutions. 

{
    \small
    \bibliographystyle{ieeenat_fullname}
    \bibliography{main}
}

\end{document}